\documentclass[10pt,twocolumn,letterpaper]{article}
\usepackage[pagenumbers]{cvpr}
\ifdefined\XeTeXversion
  \usepackage[OT1]{fontenc}
\fi

\usepackage{cuted} 

\usepackage{currfile} 

\usepackage{caption} 

\usepackage{colortbl}
\definecolor{cfoursrow}{RGB}{232,242,247}

\usepackage{graphicx}
\usepackage{amsmath}
\usepackage{amssymb}
\usepackage{booktabs}
\usepackage{tabularx}
\definecolor{cvprblue}{rgb}{0.21,0.49,0.74}
\usepackage[pagebackref,breaklinks,colorlinks,allcolors=cvprblue]{hyperref}

\newcommand{\papertitle}{Collaborative Forcing: eRank-Guided Burst Scheduling for Autoregressive Video Generation}
\author{%
\textbf{Zirui Lang}\thanks{Equal contribution.}\kern.3em$^{1}$ \quad
\textbf{Xijun Wang}\footnotemark[1]\kern.3em$^{2}$ \quad
\textbf{Wenbiao Zhao}$^{3}$ \quad
\textbf{Runzhong Qiao}$^{4}$ \quad
\textbf{Saining Zhang}\thanks{Corresponding author.}\kern.3em$^{5}$\\[3pt]
$^{1}$Beijing Institute of Technology \quad $^{2}$University of Science and Technology of China\\
$^{3}$Xinjiang University \quad $^{4}$Tsinghua University \quad $^{5}$Nanyang Technological University\\[5pt]
{\normalsize\urlstyle{same}Project Page: \url{https://langzirui3-sys.github.io/collaborative-forcing-project/}}
}
\makeatletter
\iftoggle{cvprfinal}{%
  \patchcmd{\@maketitle}{\vspace*{24pt}}{\vspace*{16pt}}{}{}%
  \patchcmd{\@maketitle}{\vspace*{12pt}}{\vspace*{-10pt}}{}{}%
}{}
\makeatother
\hypersetup{
  pdftitle={Collaborative Forcing: eRank-Guided Burst Scheduling for Autoregressive Video Generation},
  pdfauthor={Zirui Lang, Xijun Wang, Wenbiao Zhao, Runzhong Qiao, Saining Zhang},
  pdfsubject={Training-free collaboration for autoregressive video generation},
  pdfkeywords={video generation, model collaboration, effective rank, KV cache}
}

\title{\papertitle}
\begin{document}
\maketitle
\begingroup
\iftoggle{cvprfinal}{\setlength{\stripsep}{6pt plus 2pt minus 2pt}}{}
\begin{strip}
\centering
\includegraphics[width=\textwidth]{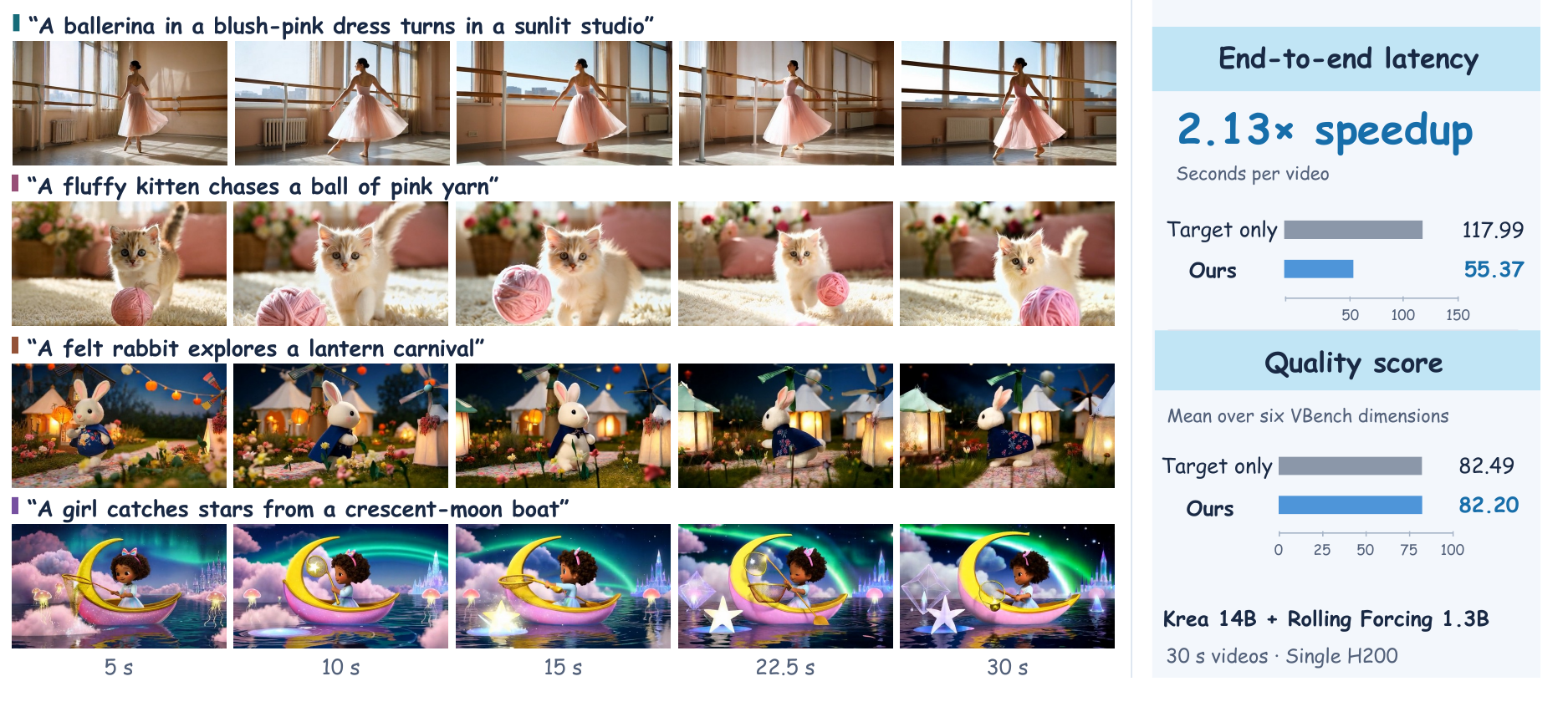}
\captionof{figure}{\textbf{Qualitative examples and end-to-end latency.}
Left: example frames sampled at 5, 10, 15, 22.5, and 30 seconds, with prompts shown above each row.
Right: a 30-second comparison of Collaborative Forcing and pure Target with Krea Realtime 14B and Rolling Forcing 1.3B on one H200. Video-ready latency drops from 117.99 to 55.37\,s ($2.13\times$).}
\label{fig:teaser}
\end{strip}
\endgroup

\begin{abstract}
Collaborative Forcing is a training-free framework that coordinates a smaller Draft and a larger Target for autoregressive video generation.
RankPulse detects relative drops in the effective rank of Draft features and schedules later Target bursts within a fixed budget.
StateBridge reuses Draft's native KV cache and incorporates missing Target outputs.
FuseStep combines that prefill with the first resumed denoising step.
On one H200 per video, we pair Krea Realtime 14B with Self-Forcing, Rolling Forcing, and LongLive at 15 and 30 seconds.
The original 128-prompt, three-seed evaluation yields \textbf{1.84--2.21$\times$} speedups over pure Target through video encoding.
Rolling Forcing at 15 seconds gains 1.35 VBench points at 1.88$\times$ speedup.
A later 384-pair LongLive 15-second rerun yields 1.79$\times$ speedup and scores 2.79 points below Target.
\end{abstract}

\section{Introduction}
\label{sec:introduction}

Autoregressive video generation has emerged as an important approach to streaming content creation and interactive world modeling~\cite{huang2025self,samuel2026fastar}. These models generate successive chunks conditioned on previously generated content. Running a large network for every chunk incurs substantial computational cost.

KV cache compression~\cite{guo2026dummy} and sparse attention~\cite{lv2026light} reduce storage and attention costs within individual models, while trajectory stitching assigns different image denoising steps to different models~\cite{pan2024tstitch}. Chunk-level collaboration complements these optimizations by letting a small model handle most of the generation while invoking a large model at selected positions. For example, SDVG~\cite{hu2026sdvg} first uses a small Draft model to generate a candidate chunk, decodes it for ImageReward~\cite{xu2023imagereward} scoring, and invokes a large Target model to regenerate rejected candidates. Candidate generation, decoding, and scoring precede each replacement. Moreover, after Target replaces a candidate, Draft continues to maintain its historical state using its own outputs, creating a mismatch between the context used for subsequent generation and the emitted video.

We therefore ask: \emph{how can small and large models coordinate generation and handoff to preserve video quality while realizing the efficiency advantages of the small model?} We investigate three related questions: when to invoke Target, how Draft should resume from Target-generated content, and how to reduce the computational overhead of handoff.

First, under a limited Target budget, fixed intervention positions cannot adapt to changing generation states, while external quality evaluation introduces additional computation. \emph{Can information already computed during Draft generation help determine when Target should take over?} VRR~\cite{chen2026vrr} links generation degradation to declining effective rank in hidden representations, while TANGO~\cite{karageorgiou2026tango} uses the model's own noise predictions to diagnose generation states. Effective rank (eRank)~\cite{roy2007effective} measures how broadly features are distributed across representational directions. It decreases as fewer directions dominate. Our diagnostics reveal that eRank and MUSIQ-SPAQ image quality~\cite{ke2021musiq,fang2020spaq} exhibit related trends after averaging across videos, while equal-budget routing controls evaluate the practical value of this signal for intervention placement. We introduce \textbf{RankPulse}, which compares the current eRank with its recent history and uses relative declines to schedule subsequent Target chunks within a fixed budget. The current Draft chunk remains committed, and routing requires neither additional candidate decoding nor an external quality evaluator.

Second, selecting an intervention position is insufficient unless Draft can properly resume from Target's outputs. Effective model collaboration requires transferring both content and state. We introduce \textbf{StateBridge}, a history-consistent handoff mechanism that synchronizes the models' conditioning histories through committed latents while allowing each model to construct its own native KV state. Before resuming generation, Draft reuses its valid historical cache and incorporates the missing Target-generated content, aligning its conditioning context with the emitted video. As an optional extension, we temporarily strengthen attention to Target history near the switching boundary to encourage its use during continuation.

Finally, history synchronization itself introduces computation. A sequential handoff first prefills Target-generated history through Draft and then executes the first denoising step of the next chunk. These operations can be evaluated jointly while respecting their causal dependencies. We introduce \textbf{FuseStep}, which combines clean-history prefill and the first continuation denoising step in a single forward pass. Block-causal attention preserves the required visibility for each chunk, and only committed history is written to the persistent cache. This design reduces separate forward calls and lowers local handoff latency.

Together, these designs form \textbf{Collaborative Forcing}, a training-free framework for collaborative autoregressive video generation. We evaluate Krea Realtime 14B~\cite{krea2025realtime} paired with Self-Forcing, Rolling Forcing, and LongLive on a single H200 GPU, using 128 prompts and three random seeds for 15- and 30-second video generation. Across these configurations, Collaborative Forcing achieves \textbf{1.84--2.21$\times$ end-to-end generation speedup} over pure Target, with quality--efficiency trade-offs that vary across model pairs.

Our main contributions are:
\begin{itemize}
    \item \textbf{An internal-representation-based online routing method.} RankPulse uses relative changes in Draft's eRank to assign subsequent video chunks under a limited Target budget, without additional candidate decoding or external quality evaluation.
    \item \textbf{A model handoff method that combines history consistency with execution efficiency.} StateBridge aligns the continuation model's native state with the accepted history, while FuseStep jointly executes history prefill and the first continuation denoising step to reduce forward calls and local handoff overhead.
    \item \textbf{An empirical evaluation of collaborative generation efficiency and quality--efficiency trade-offs.} Across three Draft backends and two video lengths on a single GPU, Collaborative Forcing achieves \textbf{1.84--2.21$\times$} end-to-end speedup over pure Target. Controlled ablations examine intervention placement, history synchronization, and joint execution.
\end{itemize}

\begin{figure*}[t]
\centering
\includegraphics[width=\textwidth]{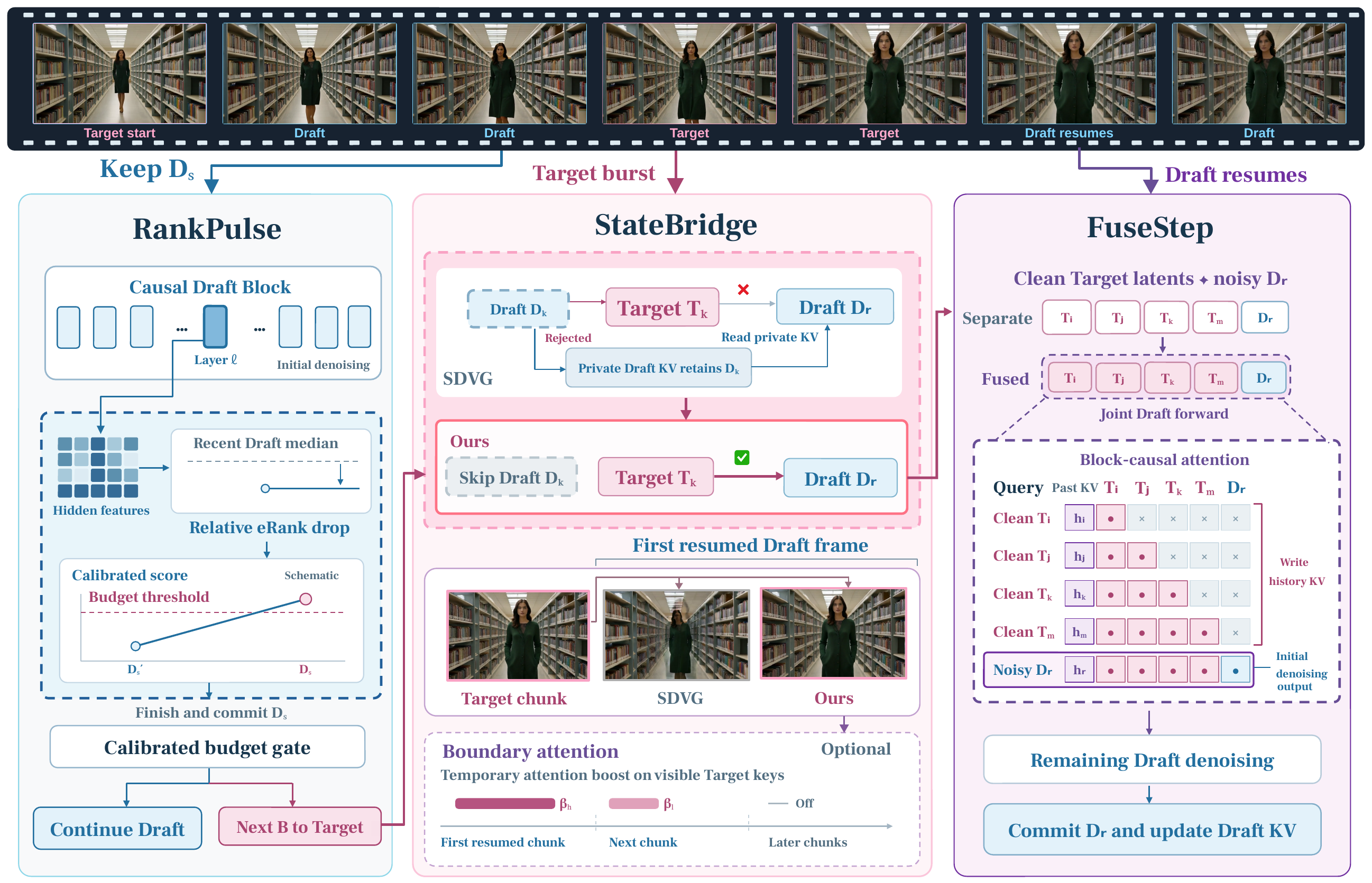}
\caption{\textbf{Collaborative Forcing on one GPU.}
The filmstrip and handoff thumbnails illustrate a fixed-schedule Self-Forcing example in a library aisle.
\textbf{RankPulse} keeps the current Draft block and uses its relative eRank drop to decide whether Target generates the next burst.
\textbf{StateBridge} extends Draft's native KV with committed Target latents; the SDVG path illustrates resumption from private Draft history.
\textbf{FuseStep} combines history prefill with the first resumed denoising step, after which Draft follows its native sampler.}
\label{fig:method}
\end{figure*}

\section{Related Work}
\paragraph{Autoregressive video diffusion.}
Diffusion Forcing~\cite{chen2024diffusionforcing} combines causal sequence modeling with per-token noise levels. CausVid~\cite{yin2024causvid} distills a bidirectional teacher into a causal generator. Causal Forcing~\cite{zhu2026causal} uses an autoregressive teacher for initialization. Self Forcing~\cite{huang2025self} trains on self-generated histories. LongLive~\cite{yang2025longlive} combines streaming long-sequence training with windowed state management. Rolling Forcing~\cite{liu2025rolling} uses joint denoising and attention sinks to support long rollouts. In-Context Forcing~\cite{yang2026incontext} varies the noise level of context frames by their distance from the current frame. Collaborative Forcing coordinates two pretrained generators at inference time while preserving their native state semantics.

\paragraph{Efficient video diffusion inference.}
TMD~\cite{nie2026tmd} distills multistep trajectories into a few transitions. FasterCache~\cite{lv2025fastercache} reuses intermediate features across denoising computations, and Pyramid Attention Broadcast~\cite{zhao2024pab} reuses attention outputs across timesteps. TeaCache~\cite{liu2025teacache} uses changes in timestep-modulated inputs to decide when to reuse computations. ARCache~\cite{nan2026arcache} guides feature reuse with historical context and applies residual correction. Speculative Forcing~\cite{choi2026speculative} skips remaining denoising steps for individual tokens accepted early. FAST-AR~\cite{samuel2026fastar} combines temporal cache compression with sparse attention, while Focused Forcing~\cite{cai2026focused} selects cached history by generated frame and attention head. For streaming video super-resolution, ReCaVSR~\cite{wang2026recavsr} reuses previously generated latents and learns how much temporal history each layer caches. These approaches reduce computation within a generator. Our routing determines which generator produces each chunk.

\paragraph{Collaboration between small and large models.}
T-Stitch~\cite{pan2024tstitch} assigns different parts of a denoising trajectory to different model sizes. DuoDiff~\cite{gallo2024duodiff} uses a shallower denoiser for early steps and a deeper one for later steps. HybridStitch~\cite{sun2026hybridstitch} selectively assigns image tokens to a larger model for refinement. In video super-resolution, RelayVSR~\cite{wang2026relayvsr} uses a large model to generate sparse keyframe references and a lightweight network to super-resolve every frame. For autoregressive video generation, SDVG~\cite{hu2026sdvg} scores a completed Draft chunk and asks Target to regenerate it if needed. Target tracks the selected chunks, while Draft continues from its own proposals. We use Draft features to schedule future chunks and update the resuming model's cache from the latents that were actually emitted.

\paragraph{Internal signals of generation drift.}
REPA~\cite{yu2024repa} aligns diffusion representations with external image features during training. CREPA~\cite{hwang2025crepa} uses neighboring-frame features to guide video-model fine-tuning. VRR~\cite{chen2026vrr} links declining eRank to drift and regularizes hidden representations during training, while TANGO~\cite{karageorgiou2026tango} uses look-ahead noise predictions for test-time adaptation. We read relative eRank drops from the existing Draft forward to allocate computation between the two models and evaluate the outputs on VBench~\cite{huang2024vbench}.

\section{Method}
\label{sec:method}

\subsection{Preliminaries}
\paragraph{Blockwise denoising.}
An autoregressive video diffusion model generates one temporal block of latent frames at a time. At position $c$, sampling starts from noise $\xi_c$ and conditions on the prompt $y$ and the committed latent history $\mathcal H_c=(z_0,\ldots,z_{c-1})$. At each step, the denoiser receives the current noisy block and its timestep. The sampler uses the prediction to construct the next step's input, so the denoising steps within a block run in sequence. The prompt and committed history stay fixed during these steps. Let $D$ denote the lightweight Draft and $T$ the larger Target. The clean block produced by model $a_c$ is
\begin{equation}
 z_c=G_{a_c}(\xi_c,y;S^{a_c}(\mathcal H_c)),\qquad a_c\in\{D,T\},
 \label{eq:producer}
\end{equation}
where $G_{a_c}$ denotes the full denoising process and $S^{a_c}(\mathcal H_c)$ is the model's conditioning state. The system commits $z_c$ at the end of sampling, giving $\mathcal H_{c+1}=(z_0,\ldots,z_c)$.

\paragraph{Causal context and KV reuse.}
In block-causal self-attention, tokens of the current block can attend to each other and to earlier blocks allowed by the model's context. With a rolling window, the model only needs the history selected by its native visibility rules. At each self-attention layer, the model reuses cached keys and values for visible history both across denoising steps and as generation advances to later blocks.

\paragraph{Prefill.}
Draft and Target share a compatible latent representation and each maintains its own native KV cache. A block produced by one model can therefore become part of the other model's context. To add a missing block, the incoming model runs the committed latent at its original position using its clean-context timestep. This prefill forward writes keys and values for that block. The model can then attend to the new entries together with its valid cached history.

\subsection{Framework Overview}
\label{sec:method-overview}
The timeline in Fig.~\ref{fig:method} shows one handoff. Target generates the first block, then Draft continues by default. The models take turns on one GPU. \textbf{RankPulse} reads Draft features to decide whether Target should generate a burst of future blocks, while keeping the current Draft block. \textbf{StateBridge} conditions the incoming model on the committed trajectory and identifies the history missing from Draft's retained cache. \textbf{FuseStep} fills that gap when Draft resumes and starts denoising the next block in the same forward.

\begin{figure}[t]
  \centering
  \includegraphics[width=\columnwidth]{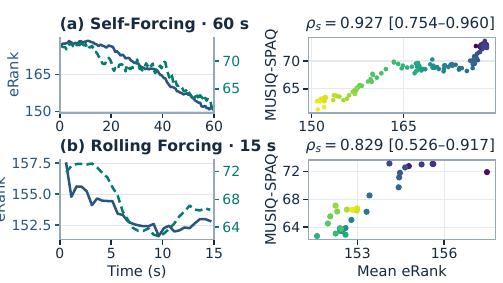}
  \caption{\textbf{eRank and image quality over generation.}
  Solid lines show mean eRank and dashed lines show mean MUSIQ-SPAQ. Dots pair block means from early (purple) to late (yellow). Brackets give 95\% prompt-bootstrap intervals for $\rho_s$.}
  \label{fig:rank-quality}
\end{figure}

\subsection{RankPulse: eRank-Guided Adaptive Routing}
\label{sec:signal}
\paragraph{Feature readout.}
In the left panel of Fig.~\ref{fig:method}, RankPulse reads visual features at Draft layer $\ell$ during the first denoising forward of the current block. It selects $N$ of $L$ visual tokens at indices $\lfloor iL/N\rfloor$ for $i=0,\ldots,N-1$. For Draft observation $s$, let $H_s\in\mathbb R^{N\times d}$ be the uncentered feature matrix. We compute effective rank~\cite{roy2007effective} from the normalized singular spectrum.
\begin{align}
 A_s &= H_s H_s^\top,\qquad \sigma_i=\sqrt{\lambda_i(A_s)},\nonumber\\
 p_i &= \frac{\sigma_i}{\sum_j\sigma_j},\qquad
 r_s=\exp\!\left(-\sum_i p_i\log p_i\right).
 \label{eq:rank}
\end{align}
The sums in Eq.~\eqref{eq:rank} use singular values from the retained positive Gram eigenvalues. A spectrum concentrated along fewer directions yields a smaller $r_s$. The readout uses an existing Draft forward and requires no frame decoding for routing.
Figure~\ref{fig:rank-quality} shows how position-averaged eRank and image quality evolve. The supplement analyzes correlations within individual trajectories.

\paragraph{Relative drop and calibration.}
Let $m_s$ be the median of up to $h$ previous Draft observations in the same trajectory. Target blocks do not advance the observation index. We set $d_1=0$ and compute subsequent drops as
\begin{equation}
 d_s=\max\!\left(0,\frac{m_s-r_s}{m_s}\right),\qquad s>1.
 \label{eq:drop}
\end{equation}
A fixed calibration set supplies reference drops $\{\delta_j\}_{j=1}^{M}$. The empirical mid-rank transform maps $d_s$ to a percentile $u_s=F(d_s)$.
\begin{equation}
 F(d)=\frac{1}{M}\sum_{j=1}^{M}
 \left(\mathbf 1[\delta_j<d]+\tfrac12\mathbf 1[\delta_j=d]\right).
 \label{eq:cdf}
\end{equation}
The percentile ranks the observed drop against the fixed calibration distribution.

\paragraph{Commit, then route.}
\label{sec:burst}
Draft finishes denoising and commits the current block before the controller selects the producer for the next position. We allocate a post-root budget of $K$ Target blocks in bursts of $B$ within a window of $P$ Draft observations. Let $b$ be the remaining bursts and $n$ the remaining eligible observations, including the current one. For $0<b<n$, the threshold is
\begin{equation}
 \theta_s=Q^{\mathrm{higher}}_{\mathcal U}(1-b/n),
 \label{eq:budget}
\end{equation}
where $\mathcal U=\{F(\delta_j)\}_{j=1}^{M}$ is the multiset of calibration percentiles. At probability $q$, $Q^{\mathrm{higher}}_{\mathcal U}$ takes the sorted entry at zero-based index $\lceil q(M-1)\rceil$. The controller triggers when $u_s\geq\theta_s$, uses the current opportunity unconditionally when $b=n>0$, and stops triggering when $b=0$.

In Fig.~\ref{fig:method}, \emph{Keep $D_s$} means the Draft block $z_c$ at observation $s$ remains committed. A trigger assigns $c+1,\ldots,c+B$ directly to Target, and Draft skips sampling those positions. Otherwise Draft generates the next block.

\subsection{StateBridge: History-Consistent State Handoff}
\label{sec:statebridge}
\begin{figure}[!t]
  \centering
  \includegraphics[width=\columnwidth]{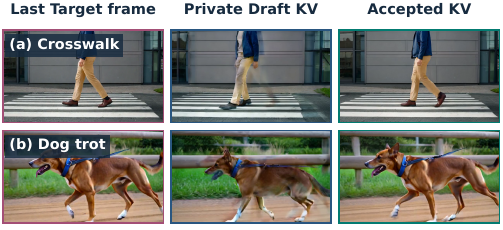}
  \caption{\textbf{Matched Target-to-Draft boundaries.}
  Each row starts with the shared last Target frame (f140), then shows the first Draft frame (f141) for each conditioning history under the same resumption noise. Cut RGB MAE is 19.9 vs.\ 8.3 for Crosswalk and 30.1 vs.\ 17.8 for Dog trot.}
  \label{fig:statebridge-matched-boundaries}
\end{figure}

A switch can leave the incoming model's cache out of step with the emitted trajectory. In SDVG~\cite{hu2026sdvg}, Target replaces a rejected Draft block in the output, while Draft's private cache retains the rejected proposal (center panel of Fig.~\ref{fig:method}). StateBridge uses committed latents for handoff. Each model's cache must correspond to the accepted content at the original positions, under that model's native context rules.
Figure~\ref{fig:statebridge-matched-boundaries} compares the first resumed Draft frame.

\paragraph{Draft to Target.}
Target builds the context it needs from the committed trajectory, including the latest Draft block, and generates and commits its burst in order. Draft retains its own native prefix KV at the switch.

\paragraph{Target to Draft.}
Let $r=c+B+1$ be the resumed Draft position and let $\mathcal Z=(z_{c+1},\ldots,z_{r-1})$ be the missing clean suffix. StateBridge reuses Draft's valid prefix KV and encodes $\mathcal Z$ at its original positions with Draft's clean-context forward. The resumed block thus conditions on the committed Target outputs. The optional boundary attention shown in Fig.~\ref{fig:method} temporarily increases the logits of visible Target keys during resumed denoising in a separate fixed-history diagnostic. The main path applies no extra logit bias.

\subsection{FuseStep: Joint Prefill and First Denoising Step}
\label{sec:execution}
\paragraph{Joint input.}
Given a valid Draft prefix cache, a sequential handoff prefills each block in $\mathcal Z$ and then runs the first denoising step for $D_r$. FuseStep concatenates the clean suffix $\mathcal Z$ and initial noise $\xi_r$ along time and evaluates them in one Draft forward. The suffix uses the native clean-context timestep, and the noisy block uses the sampler's first denoising timestep.

\paragraph{Block-causal attention.}
Each row of the attention diagram in Fig.~\ref{fig:method} represents one query block. A clean block sees its own tokens and earlier history within the native window. It cannot attend to later suffix blocks or to noisy $D_r$. The resumed block sees the eligible committed history and its own noisy tokens. Each query block keeps the same positions and visible window as in a separate forward.

\paragraph{Cache update and continuation.}
The joint forward adds only the clean suffix to persistent KV and returns the first denoising output for $D_r$. Each remaining denoising step depends on the preceding output and runs in sequence. When the block is complete, Draft writes its clean KV. Counting Draft forwards from the start of handoff through the first resumed denoising step, sequential prefill takes $B+1$ calls, batched prefill with a separate first step takes two, and FuseStep takes one. All three keep the same sampler steps and the same final clean KV write.

\section{Experiments}
\label{sec:experiments}
\begin{figure*}[!t]
  \centering
  \includegraphics[width=\textwidth]{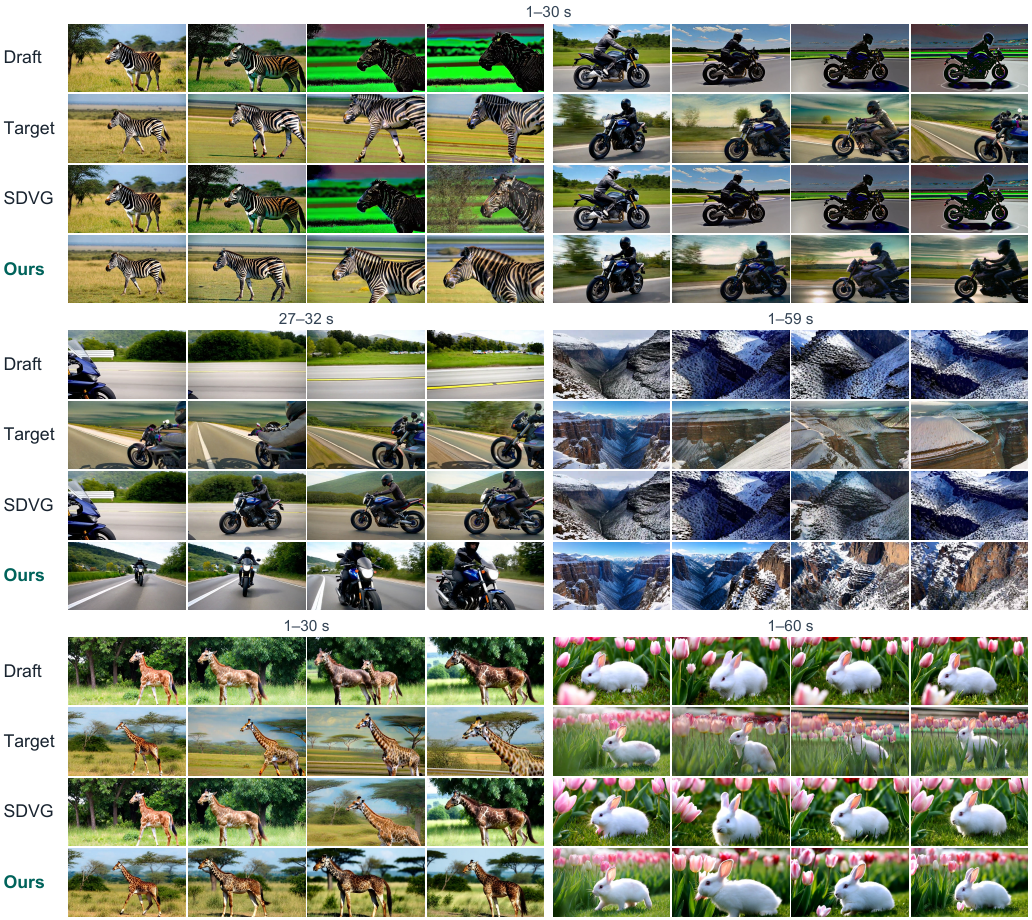}
\caption{\textbf{Qualitative comparison across Draft backends.}
\textbf{From top to bottom, the Draft backends are Self-Forcing 1.3B, Rolling Forcing 1.3B, and LongLive 1.3B, each paired with Krea Realtime 14B as Target. Each group contains two prompts.}
SDVG uses the ImageReward adaptation. Within each case, the methods share the prompt, seed, and frame times.
Ours uses one Target root plus 36 additional Target chunks for Self-Forcing and four each for Rolling Forcing and LongLive.
The Rolling Forcing motorcycle shows a selected temporal window.
\textbf{Self-Forcing (30\,s), left / right.} \emph{a zebra taking a peaceful walk} / \emph{a motorcycle accelerating to gain speed}.
\textbf{Rolling Forcing (60\,s), left / right.} \emph{a motorcycle accelerating to gain speed} / \emph{Snow rocky mountains peaks canyon. snow blanketed rocky mountains surround and shadow deep canyons. the canyons twist and bend through the high elevated mountain peaks.}
\textbf{LongLive, left (30\,s) / right (60\,s).} \emph{a giraffe taking a peaceful walk} / \emph{A small fluffy white rabbit with pale pink ears hops through a garden of blush-pink tulips and fresh green grass. It pauses to sniff a flower, gently twitches its nose and ears, then makes a few small hops forward. Soft morning daylight, delicate pink and cream tones, smooth low-angle camera tracking, cinematic realism, one continuous shot.}
The time ranges are rounded to whole seconds. The supplement lists the exact frame indices.}
  \label{fig:qualitative-model-pairs}
\end{figure*}

\begin{table*}[t]
\centering
\caption{\textbf{Single-GPU results at 15 and 30 seconds.} Each generation uses one H200. Each comparison uses 128 prompts and three random seeds per prompt. Quality is the arithmetic mean of six VBench dimensions on a 0--100 scale. Latency measures end-to-end video generation time, including decoding and encoding. Speedup is computed relative to the matched Target-only baseline within each model pair and duration. The supplement gives additional comparisons and their matched Target references. All Ours rows use FuseStep; the Rolling Forcing and LongLive Ours runs use $K=4$. Shaded rows indicate our method.}
\label{tab:main}
\small
\setlength{\tabcolsep}{2.6pt}
\begin{tabularx}{\textwidth}{@{}X*{6}{c}@{}}
\toprule
& \multicolumn{3}{c}{15-second videos} & \multicolumn{3}{c}{30-second videos} \\
\cmidrule(lr){2-4}\cmidrule(lr){5-7}
Method & Quality $\uparrow$ & Latency (s) $\downarrow$ & Speedup $\uparrow$ & Quality $\uparrow$ & Latency (s) $\downarrow$ & Speedup $\uparrow$ \\
\midrule
\multicolumn{7}{l}{\textbf{Krea Realtime 14B + Self-Forcing 1.3B}} \\
Draft only & 82.18 & 21.92 & $2.71\times$ & 77.73 & 45.25 & $2.61\times$ \\
Target only & 82.97 & 59.48 & $1.00\times$ & 81.46 & 117.98 & $1.00\times$ \\
SDVG adaptation & 82.33 & 38.46 & $1.55\times$ & 80.52 & 79.72 & $1.48\times$ \\
\rowcolor{cfoursrow}
\textbf{Ours} & 82.51 & 29.29 & $2.03\times$ & 80.71 & 63.99 & $1.84\times$ \\
\midrule
\multicolumn{7}{l}{\textbf{Krea Realtime 14B + Rolling Forcing 1.3B (Causal extension)}} \\
Draft only & 82.51 & 22.04 & $2.70\times$ & 81.64 & 45.33 & $2.60\times$ \\
Target only & 82.66 & 59.46 & $1.00\times$ & 82.49 & 117.99 & $1.00\times$ \\
SDVG adaptation & 82.57 & 49.55 & $1.20\times$ & 82.21 & 92.18 & $1.28\times$ \\
\rowcolor{cfoursrow}
\textbf{Ours} & 84.01 & 31.60 & $1.88\times$ & 82.20 & 55.37 & $2.13\times$ \\
\midrule
\multicolumn{7}{l}{\textbf{Krea Realtime 14B + LongLive 1.3B}} \\
Draft only & 77.91 & 22.36 & $2.66\times$ & 77.51 & 45.14 & $2.62\times$ \\
Target only & 82.66 & 59.44 & $1.00\times$ & 81.47 & 118.20 & $1.00\times$ \\
SDVG adaptation & 75.88 & 40.71 & $1.46\times$ & 76.92 & 79.32 & $1.49\times$ \\
\rowcolor{cfoursrow}
\textbf{Ours} & 81.11 & 29.90 & $1.99\times$ & 80.41 & 53.37 & $2.21\times$ \\
\bottomrule
\end{tabularx}
\end{table*}

\subsection{Experimental Setup}
\label{sec:setup}
\label{sec:implementation}

\paragraph{Models and baselines.}
We pair Krea Realtime 14B~\cite{krea2025realtime} as Target with Self-Forcing, Rolling Forcing, and LongLive at 1.3B. The Rolling Forcing rows use the Causal Forcing long-video extension~\cite{zhu2026causal}, which adopts Rolling Forcing's denoiser~\cite{liu2025rolling}. Baselines include native Draft, pure Target, and an ImageReward-based adaptation of SDVG~\cite{hu2026sdvg}. The adaptation uses a Target-generated first block. For later blocks, it decodes each Draft candidate and regenerates it with Target if the minimum frame reward falls below $-0.7$. Native Draft generates its own first block.

\paragraph{Evaluation protocol.}
We evaluate both nominal durations, 15 and 30 seconds, for each of the three model pairs. Table~\ref{tab:main} reports matched results on 128 prompts with three random seeds per prompt for each configuration. Draft and Target share one H200 for each main-comparison video. Matched methods use the same prompts, seed, resolution ($832\times480$), frame rate (24\,fps), and encoder settings. We give the hardware for each ablation separately.

\paragraph{Metrics.}
We average six VBench~\cite{huang2024vbench} dimensions on a 0--100 scale: subject consistency, background consistency, motion smoothness, dynamic degree, aesthetic quality, and imaging quality. We also report dynamic degree. We measure end-to-end generation latency as video-ready time, from cached text conditioning until video encoding and output have finished. It includes generation, routing, cache updates, decoding, and encoding. Model loading, warm-up, and quality scoring are outside the timer. Speedup divides the matched pure-Target mean time by the method's mean time.

\paragraph{RankPulse and Target budget.}
RankPulse reads 256 visual tokens from Draft layer 15 at the first denoising step and compares eRank with the median of up to four recent Draft observations. It retains the current Draft chunk and assigns subsequent chunks directly to Target when it triggers a burst. Each backend uses a frozen calibration set of 10 prompts. The default schedule starts with a Target block and routes one four-block burst within the first eight Draft observations. Self-Forcing at 30 seconds instead uses nine four-block bursts. The first burst follows the root. Later bursts respond to positive eRank drops with a two-observation deadline. The first two bursts use native four-step sampling. The other seven use two Target updates at the burst entry and one at each remaining block. The total is 37 Target blocks and 71 denoising updates, including the root.

\paragraph{StateBridge and FuseStep.}
StateBridge hands off committed latents and preserves each backend's native KV semantics. When Draft resumes, it reuses its valid KV cache and fills in the missing Target history. The main comparisons use FuseStep to combine that prefill with the first resumed Draft denoising step in one block-causal forward. The ablations isolate the effect of the conditioning history from the cost of processing it.

\subsection{Main Results}
\label{sec:main-results}
Across the six matched configurations in Table~\ref{tab:main}, Collaborative Forcing achieves a $1.84$--$2.21\times$ end-to-end video-ready speedup over matched pure Target. Relative to Target, Rolling Forcing gains 1.35 quality points at 15 seconds and loses 0.29 points at 30 seconds. Self-Forcing and LongLive score up to 1.55 points lower. Native Draft remains faster, while our method improves on its quality by 0.33--3.20 points.

\paragraph{Independent protocol reruns.}
A later LongLive 15-second rerun used a calibration set with no evaluation
prompt overlap and covered all 384 prompt--seed pairs on matched single H200
GPUs. The six-dimension mean was 78.62 for Ours, 81.41 for Target, and 76.97 for
Draft. Mean video-ready times were 38.53, 68.96, and 30.75 seconds, respectively,
for a $1.79\times$ speedup over Target. In a separate 32-pair Rolling Forcing
15-second shard, Ours scored 84.57 versus 82.24 for Target and reached video
output in 40.33 versus 69.21 seconds ($1.72\times$). The Rolling result covers
one shard and one seed. Both checks use a later, separately calibrated runtime
and are reported alongside the original results in Table~\ref{tab:main}.

Compared with the SDVG adaptation, Collaborative Forcing achieves a \(1.25\)--\(1.66\times\) speedup and scores higher in five of six configurations. The remaining case, 30-second Rolling Forcing, scores 82.20 versus 82.21, while latency drops from 92.18\,s to 55.37\,s.

Figure~\ref{fig:qualitative-model-pairs} compares four methods on two prompts for each Draft backend. Figure~\ref{fig:appendix-appearance} adds a matched four-method snowy-canyon example at 30 seconds. Figure~\ref{fig:appendix-archive-pairs} shows three further Draft/Ours pairs. In Self-Forcing, Draft develops strong green banding in the zebra scene, and SDVG shows the same artifact at 22\,s (Fig.~\ref{fig:qualitative-model-pairs}). In Rolling Forcing, Draft and SDVG show repeated textures on the snowy slopes. In LongLive, a second giraffe appears in Draft at 20\,s, while Ours shows one in each sampled frame. \iftoggle{cvprfinal}{%
  Table~\ref{tab:supp-main-frame-index} lists the frame indices used in the
  qualitative comparisons.%
}{%
  The supplement provides frame indices and describes the accompanying videos (Sec.~\ref{sec:supp-frame-index}).%
}

\subsection{Ablation Studies}
\label{sec:module-ablation}
\label{sec:remaining}

\paragraph{RankPulse.}
Under the same Target budget, RankPulse achieves a mean quality score of 82.55, improving over random and fixed placement by 2.98 and 2.94 points, respectively (Table~\ref{tab:routing-execution-ablation}(a)). These results support using Draft's internal signals to guide Target intervention. The supplement reports paired confidence intervals and additional placement comparisons.

\paragraph{StateBridge.}
On one H800, we fix the switch position, committed Target prefix, and resumed Draft noise. We compare resuming from private Draft history with resuming from a native cache that incorporates the accepted Target latents. Across 26 paired cases from 13 prompts and two seeds, using accepted history lowers RGB MAE at the switch from 28.16 to 10.51 and improves all 26 pairs (Fig.~\ref{fig:statebridge-matched-boundaries}). In the first resumed second, flow P90 rises from 1.51 to 4.00 pixels and the motion-compensated luma residual rises from 2.516 to 2.852.

\paragraph{FuseStep.}
On one H800, we fix the prefix cache, four clean Target blocks, and resumed Draft noise. We interleave 12 measurements per execution mode, timing handoff through the first resumed denoising step. Table~\ref{tab:fusestep-ablation} compares sequential prefill, batched prefill, and the fused forward. FuseStep saves 42.66\,ms (5.80\%) against sequential prefill and 12.54\,ms (1.78\%) against batched prefill. First-step and final latents, committed KV, cache cursors, and eRank agree.

\begin{table}[t]
\centering
\small
\setlength{\tabcolsep}{4pt}
\begin{tabularx}{\columnwidth}{@{}Xrrr@{}}
\toprule
Draft handoff & Calls $\downarrow$ & Time (ms) $\downarrow$ & GiB $\downarrow$ \\
\midrule
Sequential prefill + step & 5 & 735.24 & \textbf{19.57} \\
Batched prefill + step & 2 & 705.12 & 20.24 \\
FuseStep & \textbf{1} & \textbf{692.58} & 20.47 \\
\bottomrule
\end{tabularx}
\caption{\textbf{Fixed-history FuseStep ablation.} Local time runs from the start of handoff through the first resumed denoising step. Time and peak allocated memory are medians over 12 interleaved measurements per variant on one H800. Calls count Draft forwards over that interval. Bold marks the minimum in each column.}
\label{tab:fusestep-ablation}
\end{table}

In the H200 full-video comparison (Table~\ref{tab:routing-execution-ablation}(b)), the reported mean latency is 41.48\,s with sequential handoff and 36.67\,s with prefix KV reuse and FuseStep. The 4.81\,s difference (11.60\%, \(1.13\times\) speedup) reflects both changes in the execution configuration.

\begin{table}[t]
\centering
\small
\setlength{\tabcolsep}{4pt}
\begin{tabularx}{\columnwidth}{@{}Xr@{}}
\toprule
\multicolumn{2}{@{}l}{\textbf{(a) Routing policy}} \\
Routing policy & Quality $\uparrow$ \\
\midrule
Random & 79.58 \\
Fixed & 79.61 \\
RankPulse (eRank) & \textbf{82.55} \\
\midrule
\multicolumn{2}{@{}l}{\textbf{(b) End-to-end execution}} \\
Execution mode & Reported latency (s) $\downarrow$ \\
\midrule
Sequential handoff & 41.48 \\
Prefix KV reuse + FuseStep & \textbf{36.67} \\
\bottomrule
\end{tabularx}
\caption{\textbf{Routing and execution ablations.}
(a) On 12 Self-Forcing development prompts at 15 seconds on one H200, all policies share a Target first block, a four-block Target burst, an eight-position candidate window, StateBridge, and FuseStep. Random averages three uniform draws per prompt; Fixed uses the last eligible position.
(b) A separate full-video comparison on 12 Self-Forcing development prompts at 15 seconds uses one H200 and fixed inputs and routing. The sequential configuration re-encodes the prefix. The combined configuration reuses valid prefix KV through StateBridge and fuses prefill with the first resumed Draft step. Reported latency includes decoding and encoding.}
\label{tab:routing-execution-ablation}
\end{table}

\paragraph{Target budget.}
In the original two-H200 experiment on 24 Rolling Forcing development prompts at 15 seconds, the two- and four-block variants share the same burst start, initial history, and FuseStep execution. Increasing the burst from two to four blocks improves mean quality from 83.42 to 84.03 (paired gain 0.61 points, 95\% CI $[0.12,1.27]$) and increases average video-ready time from 28.83 to 31.67\,s. Dynamic degree is 100.00 in both variants. A later one-H200 repeat gives 83.40 and 83.98 quality points and 37.29 and 40.02 seconds for the two- and four-block variants, respectively.

\section{Limitations}
The matched 15-second H200 ablation uses 12 development prompts. The six selected VBench dimensions offer limited coverage of prompt adherence. The routing gain measured for Self-Forcing may vary across backends, and LongLive's dynamic degree remains below that of pure Target in the main comparison. Each model pair requires compatible latents and a native cache update. The runtime comparison measures video-ready latency. Energy and service throughput lie outside this measurement.

\section{Conclusion}
\label{sec:conclusion}
Collaborative Forcing uses eRank to schedule Target bursts and synchronizes Draft's cache with accepted outputs. FuseStep combines cache prefill with the first resumed denoising step. In the original single-GPU comparisons across three Draft backends, Collaborative Forcing achieves a $1.84$--$2.21\times$ video-ready speedup over matched pure-Target runs. At 30 seconds with Rolling Forcing, latency falls from 117.99 to 55.37\,s with a 0.29-point drop in mean quality.

{\small
\bibliographystyle{ieeenat_fullname}
\bibliography{./references}
}
\clearpage
\renewcommand{\thefigure}{S\arabic{figure}}
\renewcommand{\thetable}{S\arabic{table}}
\renewcommand{\theHfigure}{supp.\arabic{figure}}
\renewcommand{\theHtable}{supp.\arabic{table}}
\setcounter{figure}{0}
\setcounter{table}{0}
\captionsetup{hypcap=false}
\setlength{\textfloatsep}{12pt plus 2pt minus 2pt}
\setlength{\floatsep}{10pt plus 2pt minus 2pt}
\setlength{\columnsep}{0.25in}
\raggedbottom
\appendix
\twocolumn[{%
\begingroup
\centering
{\Large\bfseries Collaborative Forcing\par}
{\large Supplementary Material\par}
{\small\textbf{Reading guide}\quad
\hyperref[sec:appendix-per-prompt]{A. Burst placement}\quad
\hyperref[sec:appendix-quality-audit]{B. Quality and timing}\quad
\hyperref[sec:supp-complete-videos]{C. Qualitative results}\quad
\hyperref[sec:supp-erank-diagnostics]{D. eRank diagnostics}\par}
\endgroup
\smallskip
\noindent This supplement extends the system-level comparison in
Sec.~\ref{sec:experiments} with per-prompt results for burst placement at
a fixed Target budget, VBench dimension scores, and timing analyses.
It also presents qualitative comparisons and aligns two eRank measurements
with sampled video frames.
\section{Burst Placement at a Fixed Target Budget}
\label{sec:appendix-per-prompt}
\begin{center}
\small
\renewcommand{\arraystretch}{1.00}
\begin{tabular*}{\textwidth}{@{\extracolsep{\fill}}lrrrr@{}}
\toprule
Prompt ID & eRank start & Fixed start & $\Delta$ vs. Fixed & $\Delta$ vs. Random \\
\midrule
001\_subject\_consistency\_vbl0268 & 6 & 9 & +0.707 & +0.710 \\
002\_subject\_consistency\_vbl0276 & 7 & 9 & -1.067 & -0.885 \\
003\_subject\_consistency\_vbl0257 & 7 & 9 & -0.367 & +0.819 \\
009\_background\_consistency\_vbl0805 & 7 & 9 & +17.120 & +6.148 \\
010\_background\_consistency\_vbl0819 & 7 & 9 & +0.371 & +11.526 \\
011\_background\_consistency\_vbl0823 & 6 & 9 & +16.344 & +15.695 \\
017\_temporal\_flickering\_vbl0042 & 8 & 9 & +0.677 & +0.565 \\
018\_temporal\_flickering\_vbl0004 & 5 & 9 & -0.222 & +0.878 \\
019\_temporal\_flickering\_vbl0008 & 6 & 9 & -1.009 & +1.336 \\
025\_motion\_smoothness\_vbl0291 & 7 & 9 & -0.261 & -0.513 \\
026\_motion\_smoothness\_vbl0282 & 7 & 9 & +3.186 & -0.742 \\
027\_motion\_smoothness\_vbl0327 & 7 & 9 & +0.026 & +0.138 \\
\midrule
Mean paired gain & & & $+2.959$ & $+2.973$ \\
95\% prompt-bootstrap interval & & & $[-0.059,7.023]$ & $[0.405,6.208]$ \\
\bottomrule\end{tabular*}
\captionof{table}{\textbf{Per-prompt effects of burst placement.}
Starts are zero-based latent chunk indices, with the Target root at chunk 0.
Positive differences favor RankPulse on the 0--100 six-dimension VBench mean.
Three Random draws are averaged within each prompt. The bootstrap resamples
prompts 20,000 times with seed 20260911. Each video uses one H200, with
different physical GPUs across selectors.}
\label{tab:appendix-prompt-routing}
\end{center}

}]
\paragraph{Experimental design.}
Table~\ref{tab:routing-execution-ablation}(a) compares burst selectors in
aggregate. An independent 12-prompt H200 development experiment tests whether
those gains recur across prompts at a fixed Target budget. The selectors share
a Target root, one four-block burst, FuseStep, and native-cache settings. At
15 seconds and seed 20260911, RankPulse and Fixed each produce one video per
prompt, and Random produces three (60 videos total).

\paragraph{Placement and final quality.}
RankPulse beats Random on 9 of 12 prompts and Fixed on 7 of 12
(Table~\ref{tab:appendix-prompt-routing}). The mean gain over Random is 2.97
points and its prompt-bootstrap interval is above zero. Against Fixed,
prompts 009 and 011 account for most of the mean gain, and the interval
includes zero. RankPulse starts at chunks 5--8. Fixed starts at chunk 9, the
last eligible position. The gain over Fixed can reflect both earlier starts
and adaptation to the video.

\newpage
\paragraph{Block-level signal.}
To assess the signal itself, we relate eRank to MUSIQ-SPAQ within a video.
For the clean-cache layer-20 analysis in Fig.~\ref{fig:rank-quality}, the
correlation across position means is 0.927. The median within-video
correlation is 0.400 and the video-and-position residual correlation is
0.140. In a separate 16-video cohort, the layer-15 first-step readout has
median within-video Spearman correlation 0.005 with concurrent MUSIQ-SPAQ
drops (95\% CI $[-0.052,0.090]$). Both signed relative drops use the
preceding four-block median. The routing comparison above measures final
video quality after the selected burst.

\paragraph{Cohort and uncertainty.}
The 12 ablation prompts come from the 128-prompt main set. The original
10-prompt calibration shares vbl0052, vbl0283, and vbl0108 with the main set
and none with the ablation. A later calibration replaces those three and is
disjoint from both sets. Matching case-folds text and normalizes whitespace.
The intervals quantify prompt-level variation for the tested seed and devices.

\twocolumn[{%
\section{Quality and Execution Timing}
\label{sec:appendix-quality-audit}
\begin{center}
\small
\renewcommand{\arraystretch}{1.18}
\setlength{\tabcolsep}{4pt}
\begin{tabular*}{\textwidth}{@{\extracolsep{\fill}}llrrrrrrr@{}}
\toprule
Backend / duration & Method & Subject & Backgr. & Smooth. & Dynamic & Aesthetic & Imaging & Mean \\
\midrule
Rolling / 15s & Ours & 87.42 & 91.40 & 98.76 & 100.00 & 57.88 & 68.63 & 84.01 \\
Rolling / 15s & Target only & 88.53 & 91.22 & 98.57 & 87.50 & 59.46 & 70.65 & 82.66 \\
\midrule
LongLive / 15s & Ours & 92.48 & 93.99 & 99.04 & 66.67 & 61.72 & 72.76 & 81.11 \\
LongLive / 15s & Target only & 88.52 & 91.20 & 98.57 & 87.50 & 59.48 & 70.67 & 82.66 \\
\bottomrule\end{tabular*}
\captionof{table}{\textbf{Quality by dimension.}
Results are computed over 128 prompts with three seeds per prompt, using one
H200 per video. Mean is the unweighted average of the six dimension scores
and corresponds to the Quality column in Table~\ref{tab:main}.}
\label{tab:appendix-six-dimensions}
\end{center}

}]
\noindent Table~\ref{tab:appendix-six-dimensions} breaks down the 15-second
Rolling Forcing and LongLive results from Table~\ref{tab:main} into the six
VBench dimensions used to compute the reported mean quality scores.

\subsection{Quality by VBench Dimension}

\paragraph{Rolling Forcing.} Ours gains 1.35 points in the reported mean,
driven mainly by the 12.50-point improvement in dynamic degree. The subject
consistency, aesthetic quality, and imaging quality scores are lower. Subject
consistency decreases by 1.11 points.

\paragraph{LongLive.} Ours gains on five of the six dimensions. The drop in
dynamic degree from 87.50 to 66.67 outweighs those gains in the six-dimension
mean.

\newpage
\subsection{Handoff and Full-Video Timing}
\paragraph{H800 handoff with a fixed cache.} The timer starts at handoff and
ends after the first resumed Draft denoising step. Both variants begin with
the same valid Draft prefix and fill in four missing Target blocks. Sequential
prefill and resumption take 735.24\,ms. Fusing them into one forward takes
692.58\,ms. Per handoff, FuseStep saves 42.66\,ms relative to sequential
prefill and 12.54\,ms relative to batched prefill
(Table~\ref{tab:fusestep-ablation}).

\paragraph{H200 full-video latency.} Table~\ref{tab:routing-execution-ablation}(b)
reports aggregate video-ready times through decoding and encoding. The
sequential configuration reconstructs the Draft prefix cache. The combined
configuration reuses valid prefix KV and applies FuseStep. Together, these
two changes reduce the reported mean video-ready time from 41.48\,s to
36.67\,s, saving 4.81\,s per video on average. The fixed-cache H800 ablation
in Table~\ref{tab:fusestep-ablation} isolates the effect of fusion.

\twocolumn[{%
\section{Qualitative Comparisons}
\label{sec:supp-complete-videos}
\noindent\begin{minipage}{\linewidth}
\centering
\includegraphics[width=.835\linewidth]{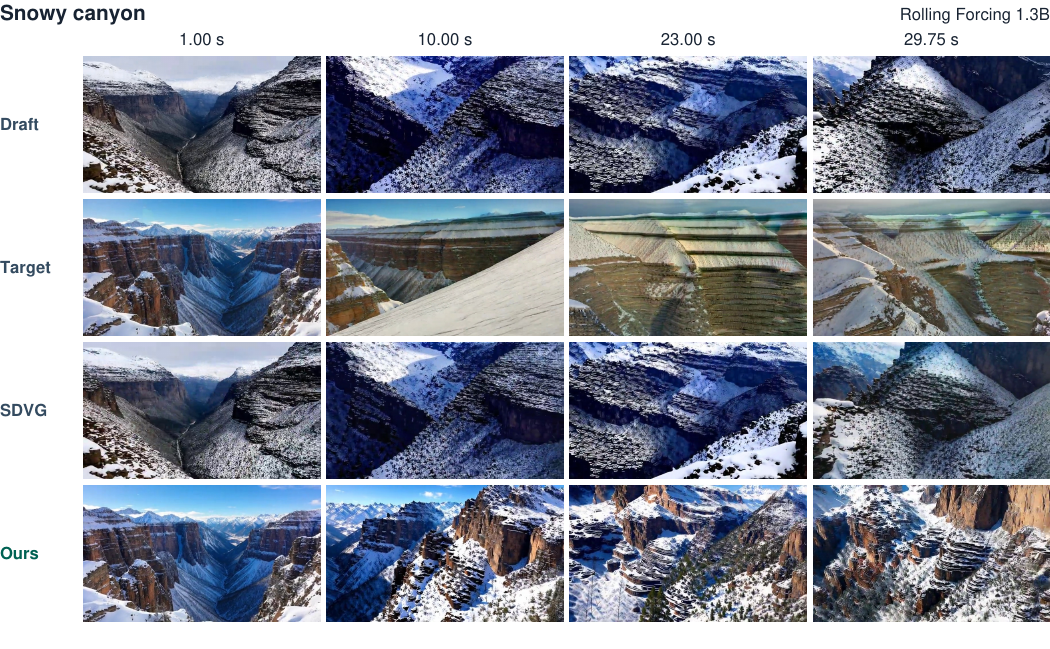}
\captionof{figure}{\textbf{Terrain detail in a 30-second snowy-canyon clip.}
All methods use the same prompt, seed 20260911, and
832$\times$480 resolution. SDVG uses the ImageReward adaptation.
Native Draft generates its own first block. Ours starts with Target.
Draft and SDVG develop repetitive slope textures, and Target develops
horizontal bands. Ours retains distinguishable rock faces and snow.
Figure~\ref{fig:qualitative-model-pairs} shows the same prompt at 60 seconds.}
\label{fig:appendix-appearance}
\end{minipage}

\medskip
\noindent\begin{minipage}{\linewidth}
\centering
\includegraphics[width=\linewidth]{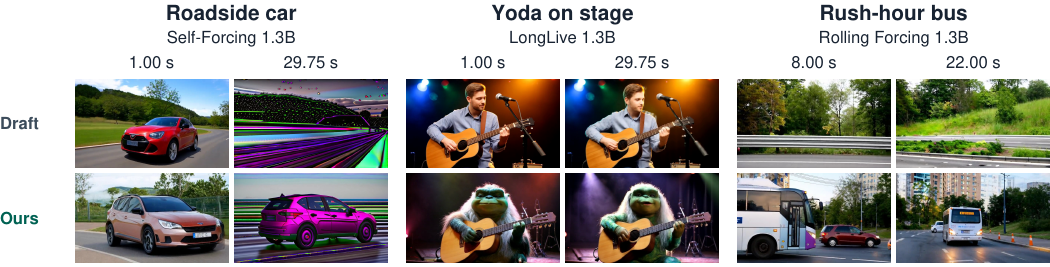}
\captionof{figure}{\textbf{Frame comparisons for three Draft/Ours pairs from development runs.}
Each pair shares its prompt, seed 20260911, resolution, and displayed times.
Ours keeps the roadside car visible despite a late magenta cast and
background bands, depicts Yoda where Draft depicts a human guitarist, and
shows more surrounding traffic in the sampled bus frames.}
\label{fig:appendix-archive-pairs}
\end{minipage}

}]
\noindent These development examples illustrate temporal changes that
cohort averages in Table~\ref{tab:main} cannot resolve. We compare complete
generation methods and separately examine the Target-to-Draft transition
with the visible Target prefix held fixed.

\subsection{Four-Method Snowy Canyon Comparison}
\noindent Figure~\ref{fig:appendix-appearance} samples the same four times
for Draft, Target, SDVG, and Ours. The frame sequence shows the progression
of repeated slope textures and horizontal banding.

\newpage
\subsection{Subject Identity and Background Detail}
\noindent Figure~\ref{fig:appendix-archive-pairs} pairs early and late
frames from Draft and Ours in three additional scenes to illustrate changes
in subject identity and background detail.

\twocolumn[{%
\begin{center}
\includegraphics[width=\textwidth]{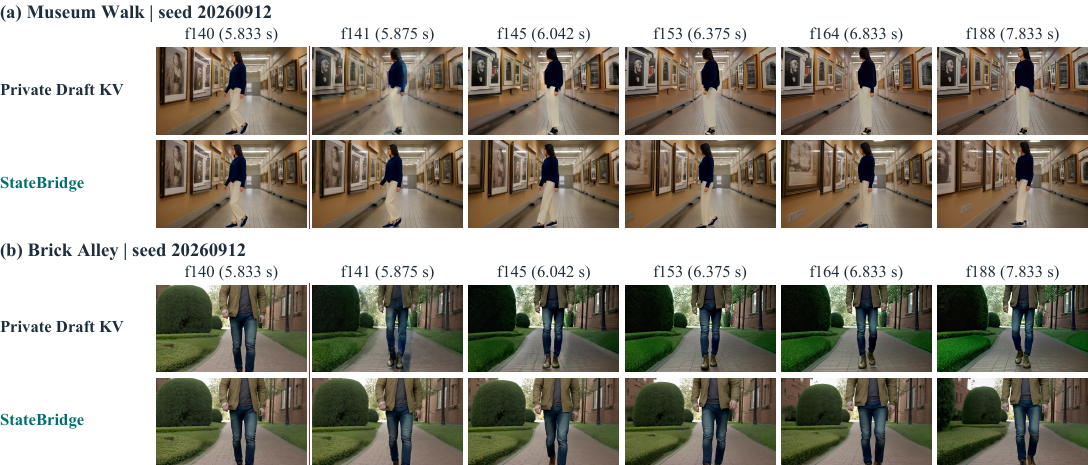}
\captionof{figure}{\textbf{Sampled frames across the Target-to-Draft switch.}
Museum Walk and Brick Alley (seed 20260912) compare Private Draft KV with
StateBridge at matched times. The red line marks the first resumed Draft
frame, f141 (5.875\,s). Indices are zero-based at 24\,fps.}
\label{fig:supp-statebridge-continuations}
\end{center}

\begin{center}
\small\renewcommand{\arraystretch}{1.10}
\newcommand{\suppframeindexrow}[4]{#1 & #2 & #3\iftoggle{cvprfinal}{}{ & #4}\\}
\iftoggle{cvprfinal}{%
  \begin{tabular*}{\textwidth}{@{\extracolsep{\fill}}llr@{}}%
}{%
  \begin{tabular*}{\textwidth}{@{\extracolsep{\fill}}llrl@{}}%
}
\toprule
\suppframeindexrow{Backend / case}{Frames}{Displayed frame indices}{Supplementary video}
\midrule
\suppframeindexrow{Self-Forcing / zebra}{717}{24, 192, 528, 716}{\textemdash}
\suppframeindexrow{Self-Forcing / motorcycle}{717}{24, 192, 528, 716}{\textemdash}
\suppframeindexrow{Rolling / motorcycle}{1437}{639, 663, 723, 759}{\textemdash}
\suppframeindexrow{Rolling / snowy canyon (60 s)}{1437}{24, 192, 720, 1416}{\textemdash}
\suppframeindexrow{LongLive / giraffe}{717}{24, 240, 480, 716}{\textemdash}
\suppframeindexrow{LongLive / garden rabbit}{1437}{24, 360, 1080, 1436}{\textemdash}
\midrule
\suppframeindexrow{Rolling / snowy canyon (30 s)}{717}{24, 240, 552, 714}{\cvprSupplementClip}
\midrule
\suppframeindexrow{Self-Forcing / roadside car}{717}{24, 714}{\textemdash}
\suppframeindexrow{LongLive / Yoda on stage}{717}{24, 714}{\textemdash}
\suppframeindexrow{Rolling / rush-hour bus}{717}{192, 528}{\textemdash}
\bottomrule
\end{tabular*}
\captionof{table}{\textbf{Frame indices for the qualitative comparisons.}
The first six rows correspond to Fig.~\ref{fig:qualitative-model-pairs},
the 30-second snowy-canyon row to Fig.~\ref{fig:appendix-appearance}%
\iftoggle{cvprfinal}{}{ and its synchronized clip}, and the final three rows to
Fig.~\ref{fig:appendix-archive-pairs}. Frame indices are zero-based at 24\,fps.}
\label{tab:supp-main-frame-index}
\end{center}

}]
\subsection{Continuity at the StateBridge Handoff}
\noindent The StateBridge ablation in Fig.~\ref{fig:statebridge-matched-boundaries}
measures the change at the first resumed Draft frame. Figure~\ref{fig:supp-statebridge-continuations}
shows subsequent frames with private Draft KV and with KV incorporating
the committed Target history. Both variants share the visible Target prefix,
handoff position, and resumed noise. The red line marks the first resumed
Draft frame.

\newpage
\iftoggle{cvprfinal}{%
  \subsection{Frame Indices}
  \label{sec:supp-frame-index}
  Table~\ref{tab:supp-main-frame-index} gives the full sequence lengths and
  the frame indices used in the qualitative comparisons. The two snowy-canyon
  entries correspond to the 30- and 60-second examples.
}{%
  \cvprSupplementPlayback
}

\twocolumn[{%
\section{Interpreting Frame-Aligned eRank}
\label{sec:supp-erank-diagnostics}
\subsection{The Runtime RankPulse Signal}
\noindent\begin{minipage}{\linewidth}
\centering
\includegraphics[width=.89\linewidth]{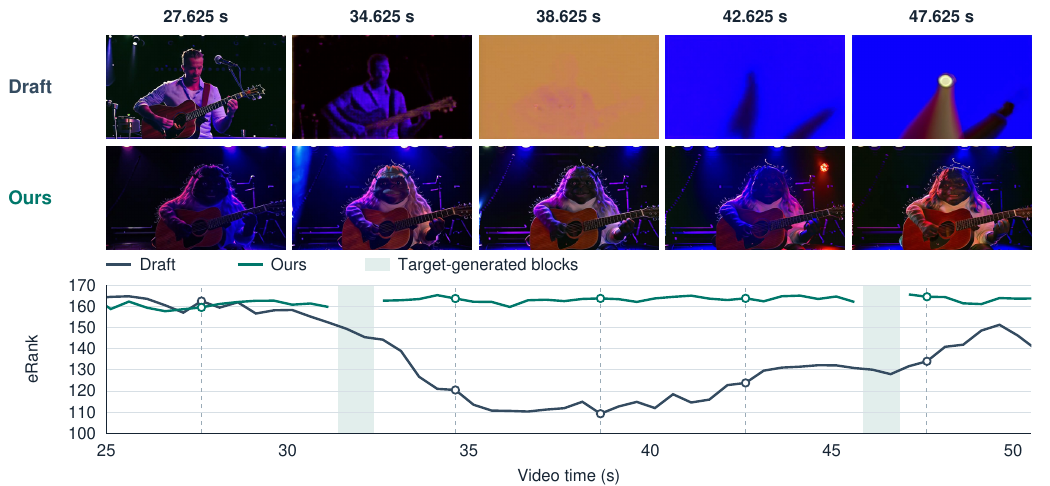}
\captionof{figure}{\textbf{Performer visibility and Draft eRank in a 60-second
Self-Forcing example.} Both methods use the Yoda guitar prompt and seed
20260911. The plot covers 25--50.5 seconds. Draft loses the performer around 38.6
seconds, while Ours retains a recognizable figure on stage. Lines show
layer-15 first-step Draft eRank. Shading marks Target-generated blocks,
where Ours has no Draft observation. Native Draft generates its own
first block. Ours starts with Target and uses eight later Target blocks
in two-block bursts.}
\label{fig:supp-rankpulse-long-case}
\end{minipage}
\subsection{Clean-Cache eRank on the Committed Video}
\noindent\begin{minipage}{\linewidth}\centering
\includegraphics[width=.89\linewidth]{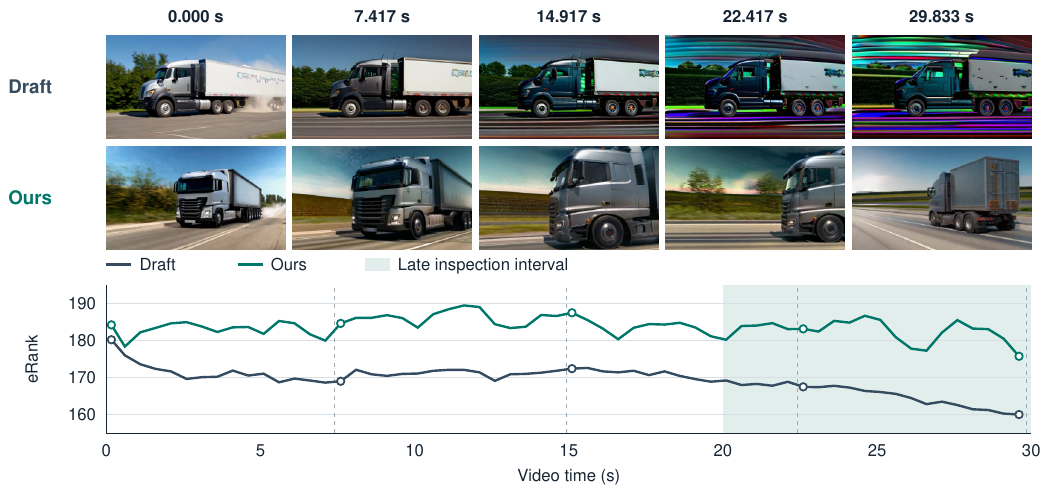}
\captionof{figure}{\textbf{Highway truck at 30 seconds.}
Horizontal bands build up around the truck in native Self-Forcing.
Collaborative Forcing retains a recognizable vehicle and road.
Both runs use seed 20260911. Dashed lines mark the displayed frame times.
Each curve connects 60 measurements of
centered layer-20 eRank from a clean forward of committed blocks at
$t=0$, including Target-produced blocks. Shading highlights 20--30 seconds
for visual inspection. Ours uses 37 Target blocks including initialization.}
\label{fig:supp-erank-clean-cases}
\end{minipage}

}]
\noindent The two eRank measurements in Sec.~\ref{sec:appendix-per-prompt}
have different scopes. RankPulse reads Draft features during generation.
\newpage
\noindent The clean-cache diagnostic reads every committed block after generation.
Dashed lines align each trace with sampled frames.
\end{document}